\documentclass{INTERSPEECH2023}

\interspeechcameraready

\title{Learning When to Speak: Latency and Quality Trade-offs for Simultaneous Speech-to-Speech Translation with Offline Models}
\name{Liam Dugan$^1$$^2$, Anshul Wadhawan$^1$, Kyle Spence$^2$, Chris Callison-Burch$^1$, Morgan McGuire$^2$, Victor Zordan$^2$}
\address{
  $^1$University of Pennsylvania, $^2$Roblox}
\email{\{ldugan, anshulw, ccb\}@seas.upenn.edu, \{kspence, vbzordan, morgan\}@roblox.com}

\begin{document}

\maketitle
 
\begin{abstract}
Recent work in speech-to-speech translation (S2ST) has focused primarily on offline settings, where the full input utterance is available before any output is given. This, however, is not reasonable in many real-world scenarios. In latency-sensitive applications, rather than waiting for the full utterance, translations should be spoken as soon as the information in the input is present.

In this work, we introduce a system for simultaneous S2ST targeting real-world use cases. Our system supports translation from 57 languages to English with tunable parameters for dynamically adjusting the latency of the output---including four policies for determining when to speak an output sequence. 
We show that these policies achieve offline-level accuracy with minimal increases in latency over a Greedy (wait-$k$) baseline.
We open-source our evaluation code and interactive test script to aid future SimulS2ST research and application development.
\end{abstract}
\noindent\textbf{Index Terms}: Simultaneous Speech-to-Speech Translation

\section{Introduction}
Speech-to-Speech Translation (S2ST) is a popular task that has seen significant recent advancements in end-to-end \cite{jia2022translatotron} and textless scenarios \cite{lee2022textless}. However, the task of adapting such models to simultaneous applications has been relatively under-explored despite the clear practical applications like cross-lingual voice chat and live interpretation.

While many such applications would benefit from simultaneous S2ST, end-to-end and textless approaches remain challenging especially for low-resource languages \cite{inaguma2022unity}.
In our work, we bypass these current research challenges by developing a robust cascaded SimulS2ST system. Our system consists of a SimulST (source speech to target text) component and a text-to-speech (TTS) component (target text to target speech). 

Inspired by Papi et al.~\cite{papi2022does}, we use an off-the-shelf offline ST model (OpenAI's Whisper \cite{radford2022robust}) and query it in an online fashion to produce accurate translations with low latency instead of training a model specifically for SimulST. We evaluate the trade-off between latency and quality for four prototyped policies for determining when to speak a given output utterance. We release our multi-threaded pipeline code and evaluations to aid future research and development in SimulS2ST\footnote{\url{http://github.com/liamdugan/speech-to-speech}}.

\section{System Design}
\begin{figure}
    \centering
    \includegraphics[width=\columnwidth]{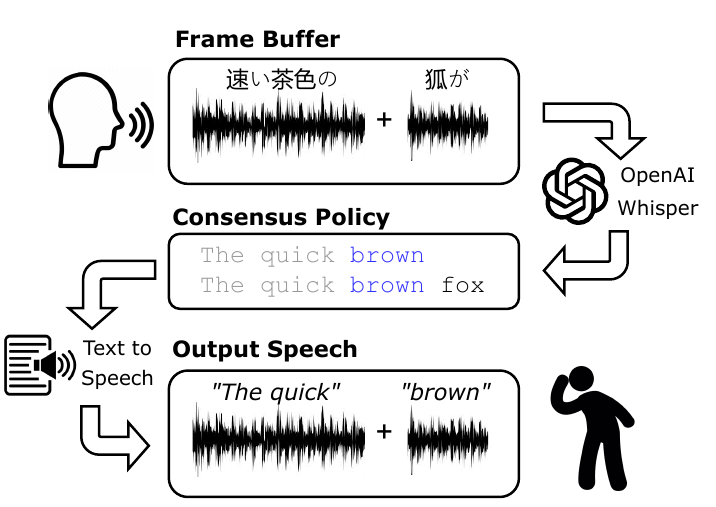}
    \caption{An example of our cascaded SimulS2ST system. Speech segments from the frame buffer are passed to Whisper. The translation is output according to the selected policy.
    }
    \label{fig:page1}
\end{figure}

In Figure~\ref{fig:page1}, we show a diagram of our system. At a high level, the system works as follows: Let $S = s_1\ldots s_N$ be the input sequence of $N$ speech frames. On each iteration we retrieve a chunk of size $w < N$ ($s_c\ldots s_{c+w}$) from the input and append it to the frame buffer $F = s_f\ldots s_{f'}, s_{f'+1} \ldots s_{f'+w}$. We then give the frame buffer and current spoken transcription $T = t_1 \ldots t_t$ as input to the ST model and generate the output text sequence $\hat{T} = t_{t+1}\ldots t_p$. This output is given to the policy $\mathcal{P}$ which determines whether or not we should speak the sequence.
If so, we use the TTS model to speak $\hat{T}$, append $\hat{T}$ to $T$, and clear the frame buffer. Otherwise, we discard $\hat{T}$ and wait for the next chunk of input speech.

We implement the SimulST and TTS model on separate threads to prevent unnecessary execution dependencies and allow for lower output latency. Additionally, when accepting live microphone input, the recorder is also given a separate thread that runs in the background during processing. Our pipeline is written in a modular fashion to allow users to quickly prototype different policies and observe their effects on the latency and accuracy of the output.

\begin{table*}
    \small
    \centering
    \begin{tabular}{l|c|c|c|c|c|c|c|c}
    \toprule
         &\multicolumn{2}{c|}{Japanese $\rightarrow$ English} & \multicolumn{2}{c|}{Spanish $\rightarrow$ English} &\multicolumn{2}{c|}{Russian $\rightarrow$ English}& \multicolumn{2}{c}{Arabic $\rightarrow$ English}\\
        Window Size ($t$)&1s&2s&1s&2s&1s&2s&1s&2s\\
        \midrule
        CAP ($\gamma = 0.9$)&20.7 (7.2)&21.1 (8.6)&38.3 (6.4)&41.5 (7.6)&36.6 (8.1)&37.4 (9.0)&18.8 (6.9)&19.5 (7.6)\\
        CAP ($\gamma = 0.5$)&15.1 (4.9)&\textbf{18.8 (6.7)}&\textbf{25.8 (3.5)}&28.2 (5.4)&28.0 (4.1)&\textbf{31.8 (5.1)}&11.7 (4.1)&\textbf{13.7 (4.7)}\\
        CP ($\alpha = 0.75$)&17.2 (6.8)&21.3 (9.6)&31.1 (4.6)&41.4 (7.9)&\textbf{27.2 (4.0)}&37.0 (8.5)&16.4 (6.2)&19.5 (8.7)\\
        CP ($\alpha = 0.5$)&\textbf{15.3 (4.8)}&20.4 (8.8)&25.3 (4.1)&\textbf{35.6 (6.1)}&21.6 (3.3)&31.3 (6.2)&\textbf{11.0 (3.7)}&19.0 (7.3)\\
        \midrule
        Greedy (wait-$k$)&5.7 (3.2)&10.0 (4.9)&8.6 (2.5)&15.2 (4.1)&12.8 (2.4)&17.3 (3.0)&3.8 (1.8)&5.9 (3.5)\\
        Offline Policy&21.9 (9.5)&21.9 (9.5)&42.9 (9.6)&42.9 (9.6)&36.6 (10.0)&36.6 (10.0)&19.7 (9.1)&19.7 (9.1)\\
        \bottomrule
    \end{tabular}
    \caption{Performance scores - BLEU (Average Lagging in seconds) - for the four policies: Offline, Greedy, Confidence-Aware (CAP), and Consensus (CP) using Whisper Medium (769M params) \cite{radford2022robust}. Bolded numbers represent optimal latency-quality trade-off. While there is no consistent winner between CAP and CP, we see that languages more structurally similar to English consistently achieve better latency-quality trade-off. In particular, Spanish and Russian BLEU scores achieve near-Offline levels with minimal increases in latency over the Greedy policy.
}
    \label{tab:online}
\end{table*}

\section{System Evaluation}
We evaluate our system on translation into English from four typologically diverse languages (Japanese, Spanish, Russian, and Arabic). For our offline ST model we use OpenAI's Whisper \cite{radford2022robust} and for our TTS model we query the ElevenLabs API \cite{elevenlabs}. We conduct our evaluations on a single NVIDIA RTX 2080 GPU and report metrics on a filtered subset of 75 examples of length 6 seconds or more from the dev set of CoVoST2 \cite{wang2020covost}.

For accuracy, we compute the BLEU score between the spoken transcript $T$ and the reference using the SacreBLEU package~\cite{post-2018-call}. We opt not to use ASR BLEU on the output speech due to the lack of precision caused by cascading ASR errors.

For our latency metric, we use Computation-Aware Average Lagging ($\text{AL}_{CA}$) \cite{ma2020simulmt} which is an adaptation of Average Lagging that takes into account the computational cost of generating an output. This is especially important for us given that we also bear a computational cost when speaking output and context switching between threads.

\subsection{Policy Comparisons}
We compare four simple policies for determining when to speak outputs. The first two are the baseline policies: the \textbf{Greedy Policy}\footnote{This is equivalent to the wait-$k$ policy from SimulST literature \cite{ma2020simulmt}} ($\mathcal{P}(\hat{T}) = \text{True}$) and \textbf{Offline Policy} ($\mathcal{P}(\hat{T}) = \text{False}$). These represent two ends of the latency-quality spectrum and help us gauge where the upper and lower bounds are for our system's quality and latency.

Next is the \textbf{Confidence-Aware Policy (CAP)}, where we return true if the average probability of the sequence returned by the ST model (i.e. the confidence of the model) is above some threshold $\gamma$. In our testing, we found that Whisper's \texttt{no\_speech\_prob} field gave better empirical results than the \texttt{avg\_logprob} field, so we use that in our evaluation.

Finally, the \textbf{Consensus Policy (CP)} returns true only when the current transcript $\hat{T}_{k+1}$ and the previous transcript $\hat{T}_{k}$'s ratio of length to edit distance is under some threshold $\alpha$. We run both this policy and the previous policy for two different parameter settings to show the capability of our system to adapt to different latency regimes.

\section{Results}
In Table \ref{tab:online} we report the results of our evaluation and find several surprising trends. First, we observe that languages more structurally similar to English tend to achieve lower latencies across all policies. In particular, we see that translations from Spanish are spoken almost a full second faster than translations from Japanese no matter the choice of policy.

Second, using our proposed policies we achieve substantial improvements in translation quality over our Greedy baseline while only sacrificing minimal extra latency.
For example, translating Spanish to English using the Confidence-Aware Policy ($\alpha = 0.5$) achieves a 17-point improvement in BLEU score over wait-$k$ with only 1 extra second of average lagging.
Likewise in Russian, using the Consensus Policy ($\gamma = 0.75$) increases the BLEU score by nearly 15 points over wait-$k$ while increasing the latency by just 1.6 seconds.

Finally, we observe that no single policy performs the best across all languages. Thus practitioners must tune their systems on a per-language basis for optimal latency-quality tradeoffs.

\section{Conclusion}
While SimulS2ST is still an active area of exploration, it already has significant practical utility for enhancing communication.
We provide a customizable baseline system for this task that allows users to dynamically tune policy parameters, directly influencing the latency-quality trade-off of their system.
Our evaluations show that these policy parameters achieve comparable accuracy to offline models while substantially improving latency over a wait-$k$ baseline. We hope our system assists industry professionals and researchers alike in developing, benchmarking, and prototyping future SimulS2ST systems.

\bibliographystyle{IEEEtran}
\bibliography{paper}

\begin{thebibliography}{1}
\providecommand{\url}[1]{#1}
\csname url@samestyle\endcsname
\providecommand{\newblock}{\relax}
\providecommand{\bibinfo}[2]{#2}
\providecommand{\BIBentrySTDinterwordspacing}{\spaceskip=0pt\relax}
\providecommand{\BIBentryALTinterwordstretchfactor}{4}
\providecommand{\BIBentryALTinterwordspacing}{\spaceskip=\fontdimen2\font plus
\BIBentryALTinterwordstretchfactor\fontdimen3\font minus
  \fontdimen4\font\relax}
\providecommand{\BIBforeignlanguage}[2]{{%
\expandafter\ifx\csname l@#1\endcsname\relax
\typeout{** WARNING: IEEEtran.bst: No hyphenation pattern has been}%
\typeout{** loaded for the language `#1'. Using the pattern for}%
\typeout{** the default language instead.}%
\else
\language=\csname l@#1\endcsname
\fi
#2}}
\providecommand{\BIBdecl}{\relax}
\BIBdecl

\bibitem{jia2022translatotron}
Y.~Jia, M.~T. Ramanovich, T.~Remez, and R.~Pomerantz, ``Translatotron 2:
  High-quality direct speech-to-speech translation with voice preservation,''
  in \emph{International Conference on Machine Learning}.\hskip 1em plus 0.5em
  minus 0.4em\relax PMLR, 2022, pp. 10\,120--10\,134.

\bibitem{lee2022textless}
A.~Lee, H.~Gong, P.-A. Duquenne, H.~Schwenk, P.-J. Chen, C.~Wang, S.~Popuri,
  Y.~Adi, J.~Pino, J.~Gu, and W.-N. Hsu, ``Textless speech-to-speech
  translation on real data,'' 2022.

\bibitem{inaguma2022unity}
H.~Inaguma, S.~Popuri, I.~Kulikov, P.-J. Chen, C.~Wang, Y.-A. Chung, Y.~Tang,
  A.~Lee, S.~Watanabe, and J.~Pino, ``Unity: Two-pass direct speech-to-speech
  translation with discrete units,'' 2022.

\bibitem{papi2022does}
S.~Papi, M.~Gaido, M.~Negri, and M.~Turchi, ``Does simultaneous speech
  translation need simultaneous models?'' \emph{arXiv preprint
  arXiv:2204.03783}, 2022.

\bibitem{radford2022robust}
A.~Radford, J.~W. Kim, T.~Xu, G.~Brockman, C.~McLeavey, and I.~Sutskever,
  ``Robust speech recognition via large-scale weak supervision,'' 2022.

\bibitem{elevenlabs}
{ElevenLabs}, ``Elevenlabs {API},'' https://api.elevenlabs.io/docs, 2023,
  accessed: 2023-04-10.

\bibitem{wang2020covost}
C.~Wang, A.~Wu, and J.~Pino, ``Covost 2 and massively multilingual
  speech-to-text translation,'' 2020.

\bibitem{post-2018-call}
\BIBentryALTinterwordspacing
M.~Post, ``A call for clarity in reporting {BLEU} scores,'' in
  \emph{Proceedings of the Third Conference on Machine Translation: Research
  Papers}.\hskip 1em plus 0.5em minus 0.4em\relax Belgium, Brussels:
  Association for Computational Linguistics, Oct. 2018, pp. 186--191. [Online].
  Available: \url{https://www.aclweb.org/anthology/W18-6319}
\BIBentrySTDinterwordspacing

\bibitem{ma2020simulmt}
X.~Ma, J.~Pino, and P.~Koehn, ``Simulmt to simulst: Adapting simultaneous text
  translation to end-to-end simultaneous speech translation,'' \emph{arXiv
  preprint arXiv:2011.02048}, 2020.

\end{thebibliography}

\end{document}